\documentclass[10pt]{article}
\usepackage{mathptmx}
\usepackage{url}
\usepackage{latexsym}
\usepackage{hyperref}
\usepackage{graphicx}
\usepackage{subcaption}
\usepackage[left=1.5in,right=1.5in,top=1in,bottom=1.4in,letterpaper]{geometry}
\usepackage{enumitem}

\title{\bf{Automatic Procurement Fraud Detection with Machine Learning}}

\author{Jin Bai, Tong Qiu}

\date{}

\begin{document}
\maketitle

\begin{abstract}
Although procurement fraud is always a critical problem in almost every free market, audit departments still have a strong reliance on reporting from informed sources when detecting them. With our generous cooperator, SF Express, sharing the access to the database related with procurements took place from 2015 to 2017 in their company, our team studies how machine learning techniques could help with the audition of one of the most profound crime among current chinese market, namely procurement frauds. By representing each procurement event as 9 specific features, we construct neural network models to identify suspicious procurements and classify their fraud types. Through testing our models over 50000 samples collected from the procurement database, we have proven that such models -- despite having space for improvements -- are useful in detecting procurement frauds. 
\end{abstract}

\section*{Keywords}

auditing, procurement fraud, machine learning, artificial neural network

\section{Introduction}

Procurement fraud, sometimes called contract fraud, is believed by professionals to be one of the most common and costly white-collar crime. \cite{anatomy} It is defined to be an intentional act by one or more individuals among management, those charged with governance, employees, or third parties, involving the use of deception to obtain an unjust or illegal advantage. \cite{240}

Fraudulent process in procurement has been bothering various corporations all over the world, including governmental departments \cite{airforce} for a long time. Problems such as collusion between bidders and bid inviters, buyers’ acceptance of bribes and creations of fictitious transactions all contain great possibilities of causing both financial and assets damages to the corporations who desire purchases of products from suppliers in the most economical way.

 As stated by Ying, vice president of Higher Education Forensic Accounting Professional Core Course Textbook Editorial Board, despite the ubiquity and profoundness of procurement fraud, the main auditing methods used by most audit departments are still book audit and reports reviewing. \cite{ping} During our contact with SF Express, such facts were admitted by the manager of auditing department of SF Express as well. In other words, even being the greatest express company in China, SF Express still lack the ability to execute initiative fraud auditing, not to mention other corporations who do not have equal scales as SF Express. After detailed communication with SF auditing department, we arrived at the conclusion that such currently unavoidable dependence on informers is mainly caused by both the huge amount of procurements operated in the company annually and the complicated steps needed in the procurement process, including bidding, enquiry, contract management and order management. Apparently, such enormous amount and complication are making the department unable to carefully audit procurements one by one with mere human power.

Considering the need for handling data of large scale and relationship between different steps and elements in a complete procurement process, we figured out that a machine learning model established with artificial neural network algorithms could be a handful tool. We decided to rely on computers' great ability of process large-scale data and take advantages of specialties of a artificial neural network including its ability to perform nonlinear computing to deal with the complex relationships between different steps and elements. We expected our model to have the ability to inform its users the probability that one procurement involves fraudulence after a set of data related to that procurement was inputted.

\section{Background}

\subsection{History of Machine Learning}

Machine learning refers to the study of giving computers ``the ability to learn without being explicitly programmed''. \cite{mlorigin} Growing from the field of artificial intellegence, machine learning researches gradually shifted their focus from serving as a tool for AI to solving practical problems using models from probability theory and statistics. \cite{mlshift} Main types of problems for machine learning include classification, regression, clustering and estimation, which relate very well with our objectives of procurement fraud detections. 

Neural networks arise from a branch of machine learning called deep learning, which contains multiple hidden layers in the learning models -- in contrast to only one layer in traditional (shallow) models. Compare to the shallow ones, deep learning models are more capable to handle handle areas superior to human insights, but also require more computing power in general. \cite{imagenet} The invention of Nvidia's CUDA framework on its Graphical Processing Units (GPUs) in late 2000s resulted in the ``big bang'' of deep learning, as it continues to become a very hot and trending subject in an enormous number of disciplines. \cite{nvidia}

\subsection{Prior Work on Machine Learning}

Various efforts have already been made to take advantage of machine learning, especially neural networks, in the area of fraud detection. The most typical case is the detection of frauds in credit card transactions. A fair number of scholarly articles have supported the idea that a neural network is a feasible approach in credit card fraud detection. \cite{creditcardnn} In addition, Kaggle, a world-famous data science website, recently hosted a machine learning challenge in exactly this area. \cite{kaggle} Based on a pre-collected, pre-labelled dataset of transactions, participants in this challenge are required to come up with the best machine learning model to identify frauds as accurately as possible.

These articles and events can benefit us a lot when moving this idea into the area of procurement auditing. They not only build up our confidence that our approach has a high chance to succeed in auditing, but also provide us with good neural network models or other learning algorithms as a starting point. On the other hand, since the Kaggle challenge relies on a detailed and organized dataset, and since the purpose of the challenge is to find the best machine learning model, we should always be aware of the two key components in this study -- reliable and comprehensive data and a carefully structured model.

\subsection{Prior Work on Procurement Fraud Detection}

As pointed out by Hugo, Badenhorst-Weiss and Van Rooyen, compared with researches on other business functions, there has been a shortage of study on procurement fraud from the risk management perspective.\cite{riskmana} Moreover, the power of machine learning has not yet been applied to procurement fraud detection and prediction at the time this study is conducted.

How to actively and effectively detect procurement has been a problem bothering enterprises and governments for a long time. Previous studies of procurement fraud investigation stated that purely manual investigation has an overwhelmingly high demand of well-trained investigators that could be unaffordable for even government departments. \cite{investigation}

Prior researchers have also built risk management models for procurement frauds and successfully proved these models to be effective. \cite{riskmodel} However, they still fail to put auditors on the aggressive side of investigation.

Last but not least, the study group designs the research to be Chinese-market-oriented due to the fact that all data we use is from a Chinese enterprise. We base our understanding of procurement fraud on Chinese laws \cite{chineselaw} as well as studies of procurement auditing by other Chinese scholars. \cite{wang}\cite{dong}

\section{Data Specification}

\subsection{Data Collection}

As we are expecting a machine learning model to be used in procurement auditing, a database containing historical records of real procurements conducted in one company is indeed necessary for machine training. Therefore, in order to make sure our machine could be “well trained”, our first step is to collect data and establish a database with both huge enough scale and a balanced ratio between the amounts of positive cases (procurements that involve fraudulence) and negative cases (procurements that do not involve fraudulence).

Being our cooperator, SF Express provides us with access to records of all their previously audited procurements dated after January 1st, 2015, as well as all procurements take place after January 1st, 2015, whose relevant information is stored in the systems applications and products database (SAP database). 

Considering the extreme asymmetry between the number of positive and negative cases, the research team decides to alleviate such asymmetry by conducting sampling on positive cases. As we have around 25000 negative cases in total, we also pulled 25000 positive cases out of the SAP database, to ensure a $1$ to $1$ ratio. When sampling positive cases, we first choose 5 random dates between 2015.1.1 and 2017.5.31 and make sure none of the 25000 fraudulence procurements take place on these dates. We then manually collect 5000 procurements from all procurements happened on each 5 dates and summarize them into one sheet using Arbutus Analyzer.

For all variable listed below(see \emph{3.3 Input Variables}), we directly used the raw data stored in SF's SAP system in order to test the ability of the machine to function efficiently in real business and to test its multiusability in different businesses. However, data inputted will be further normalized during the process of machine learning.(see \emph{5.1 Implementation Details})

\subsection{Choice of Variables}
As SF Express only provides us with the access to the user interface of SAP system, other than the admin zone of it, we have no choice but to limit the number of variables used in machine learning. The research team decide to choose input variables which satisfy the criterions listed by Robert May, Graeme Dandy and Holger Maier \cite{inputs}. That is, all variables must be chosen with the following five factors considered:

\subsubsection*{\emph{Relevance}} Apparently, all input variables of the model must contain their own relevance with the output. In our research, input variables need to have at least one feature that could be the direct cause of fraud or an obvious clue of fraud detection. Nevertheless, we also predict some variables might not be proved to have such required features as they were previously studied by linear models even though non-linear models as Neural Network may actually discover such features of these variables.

\subsubsection*{\emph{Computational Effort}} With the increase of number of input variable, the server will have to bear more computational burden. The most direct effect of such burden would be the substantial expansion of computing time. In pratice, considering the concrete use of our model, efficiency should always be a necessary trait as the model is designed mainly for business use. After all, there is no point of using a machine learing model if it is even less efficient than  
human censoring.

\subsubsection*{\emph{Training Difficulty}} Another problem that could be caused by the increasing number of variables is the difficulty for Nueral Network builders to sufficiently train their models. Redundant and irrelevant variables can slower the training speed since they increase the number of possible combinations of parameters, which will create locally optimal error values. Also, the fact that such variables are redundant and irrelevant can result in longer time for the machine to recognize their relationship with the error and to successfully map such ambiguous relationship. Unfortunately, in our study, redundant and irrelevant facts about procurements occupy a great amount of the whole SF database and thus result in the very limited number of variables eventually put into study. This problem will be further discussed in our expectation of future studies.

\subsubsection*{\emph{Dimensionality}} A critical fact that one should learn about artificial neural network is the \emph{curse of dimensionality} \cite{bellman}, the fact that as the dimensianlity of the model increases linearly, the total volumn of the domain of the modelling problem would increase exponentially. As the completion of one procurement requires step by step operation that involves the supplier, the purchaser and the regulatory authorities, the establishment of a multi-dimensional model is indeed unavoidable. In the study, our team decides to establish a total of four dimensions for the training machine, including: 1. numerical data of the procurement 2. information about the supplier 3. information about the purchaser 4. the property of fraud.

\subsubsection*{\emph{Comprehensibility}} While early researchers and modellers like to refer to artificial neural network as a "black box" \cite{blackbox}, recent studies ask models to be more self-explanatory. In particular, the fulfillments of the following 3 purposes are required:

\begin{enumerate}[
    wide=0pt,
    noitemsep,
    labelwidth=2em,
    labelsep*=1pt,
    itemindent=0pt,
    leftmargin=\dimexpr\labelwidth+\labelsep\relax,
    itemsep=10pt,
    label=\roman*.
]
\item The inputs should have a certain domain that produce certain outputs, which can be useful knowledge in the neural network itself.
\item The model should be able to verify that the response trends between the input and output data make sense.
\item The model should be able to discover new relationships between inputs and outputs, which reveal previously unknown insights into the underlying physical process. \cite{C&S}
\end{enumerate}

\subsection{Input Variables}

\begin{enumerate}[
    wide=0pt,
    noitemsep,
    labelwidth=2em,
    labelsep*=1pt,
    itemindent=0pt,
    leftmargin=\dimexpr\labelwidth+\labelsep\relax,
    itemsep=10pt,
    label=\roman*.
]
\item \textbf{Procurement Serial Number (PSN)}: This is the serial number used in SAP system to identify each procurement. With each serial number, details of individual procurement could be traced. We choose serial number as a variable because when series of fraud take place, they could have similar or consecutive serial numbers. Another reason to choose this variable is that similar or consecutive serial numbers are capable of leading auditors to procurements which belongs to the same procurement contract or procurement program.

\item \textbf{Procurement Group Number (PGN)}: The very basic logic to choose procurement group number as an input variable is that crime and deviance have the likelihood to repeat in specific circumstances or neighborhoods. \cite{criminology} With the memory function of machine learning, we expect the machine to conclude which type of procurement group would have a higher chance of committing fraud. On the other hand, superior's negligence is a cause of fraud. Therefore, we assume a specific procurement group can have more possibility to conduct fraud than another group.

\item \textbf{Procurement Organization Number (PON)}: Procurement organization number as a variable is chosen for almost the same reason as procurement group number. In previous procurement fraud studies, researches proved the importance of management in fraud prevention. \cite{management} Unlike procurement group number, which refers more to a specific group who organize the procurement in detail, procurement organization number is rather a representation of the manager who leads the groups.

\item \textbf{Material Group Number (MGN)}: We decide to include this input variable based on the assumption that certain types of fraud like receiving kickbacks are more likely to happen during the purchase of specific types of products. What's more, the material group that a product belongs to also contains information of both the product itself and its possible compliments and substitutes. Such information could be helpful in further building of the model when computations of price elasticities are used to judge the legitimacy of one procurement.

\item \textbf{Net Price (NP)}: People's intention to conduct procurement fraud is related with the profit they could get from such action, which is then related to the net price of the product being purchased. In early studies of fraud auditing, net price is proved to have direct relationship with the occurence of fraud as well as the shrinkage of a company's economic benefit. \cite{bijia}

\item \textbf{Purchase Amount (PA)}: One forgettable fact that is related to procurement fraud is that fraudsters can gain a large amount of profit even when his (or her) profit per unit is negligible. Especially in large scale companies who execute a substantial amount of purchases monthly, a fraudster gaining a little profit per unit can result in huge losses for the company. Also, according to experiences of workers of SF auditing department, smallness of unit profit usually increases the difficulty of manual auditing. However, that would surely not bother a model who has hundreds of millions of times computing power.

\item \textbf{Procurement Total Price (PTP)}: This is an inclusive representation of the above two factors, which more generally reflect the size of one business. It will not only provide the machine a detailed insight of how scale of business is related with possibility of fraud but also give the machine a better idea on the classification of each case. (See below, \emph{Fraud Type})

\item \textbf{Fraud Type (FT; If no fraud is contained, this value would be set to 0)}: We train the machine with fraud cases that are already classified so that it would have the ability to predict not only the possibility of fraud but also what kind of fraud one specific case might belong to. In practice, this feature would make auditors more efficient when they perform more detailed examination of a procurement at a later time. Also, this variable would help the machine discover special relationships between one type of fraud and other parameters, which exactly meets the need of the 3rd standard of comprehensiveness listed in \emph{Choice of Variables}.

\item \textbf{Supplier Serial Number (SSN)}: Previous studies on procurement fraud have found fraud undetectable because it usually involves collusion between members of staff and suppliers. \cite{supplier} With supplier being one important part (as both conductor and benefiter) of conducting procurement, we expect suppliers who have a crime history to have more chances to break the law again. In the logic of machine learning, we feed the machine with suppliers who have crime histories so that it would automatically build a blacklist which makes the machine more "cautious" when same suppliers appear again.
\end{enumerate}

Below is a sample input chart to help readers better understand what our input actually look like.

\begin{figure}[h!]
    \centering
    \includegraphics[width=0.8\textwidth]{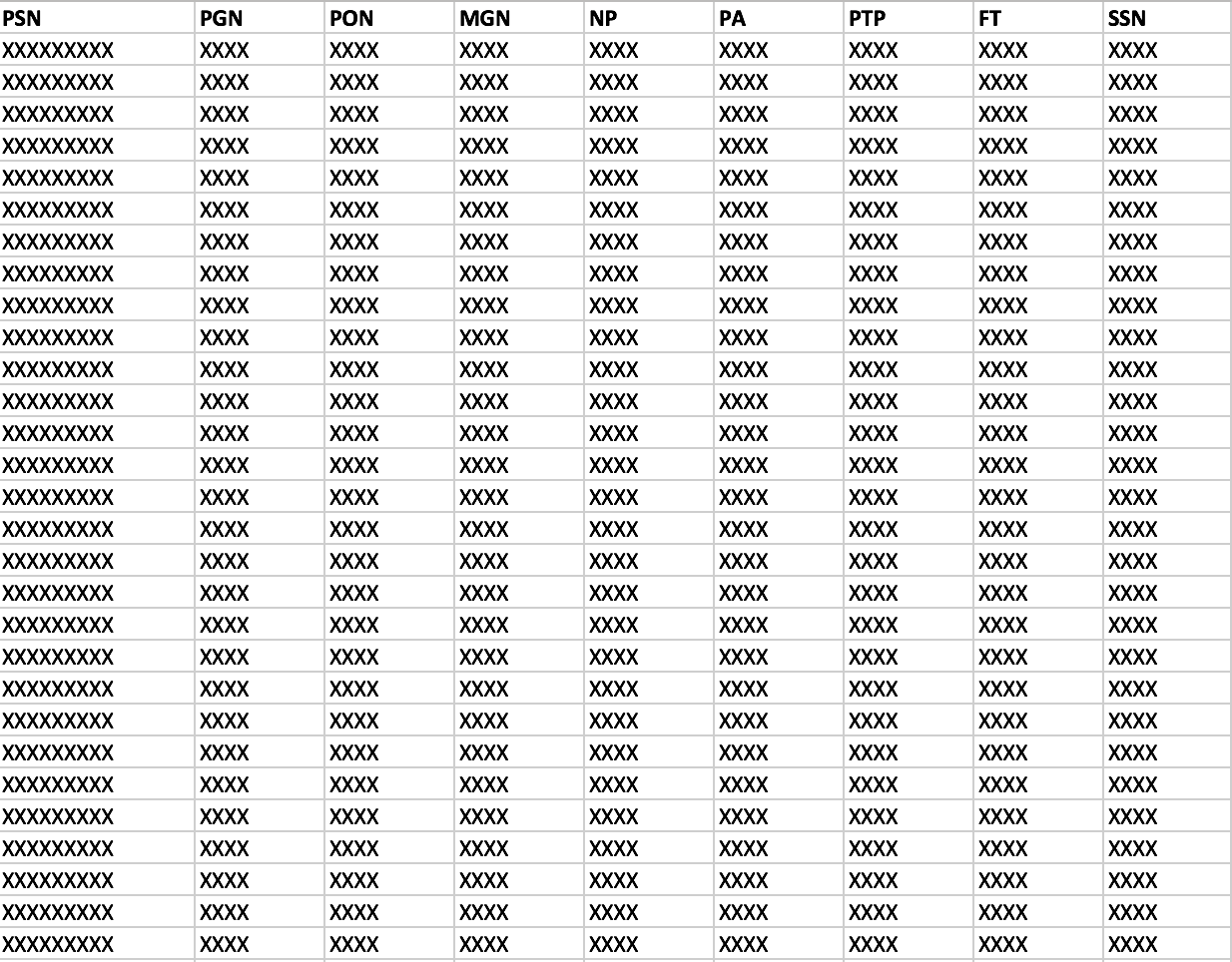}
    \caption{Input sample}
    \label{fig:sample}
\end{figure}

\section{Neural Network Model}

\subsection{From Basic Concepts to MLP}

A neural network consists of several layers that manipulate on multi-dimensional arrays of data, often called tensors. \cite{tensor} Each layer can be viewed as a function, whose both input and output are tensors of particular shapes. Internally, it has a set of parameters, which, given an input, compute the output based on a generalized version of matrix multiplication. Multiple such layers are stacked together, so that an input tensor of data is ``fed forward`` through these layers to produce the output, which is distinct enough for a machine to generate different predictions from different inputs.

The training of a neural network is, therefore, a process to improve the parameters of the layers, so that the output could lead to the right prediction of a given task. Given an input to the model, the prediction generated by the output of the network is compared to the ``label'' -- the ground truth corresponding to that input. If the results do not coincide, a ``loss function'' is calculated and applied to every layer to compute its ``gradient'' -- the optimal direction to update this layer so it can generate outputs for more accurate predictions. The layer is then slightly adjusted along the gradient direction, a process called ``gradient descent''. \cite{gd} The above process is repeated thousands or millions of times, until the network reaches its optimum, where, given any input, the model has a highest chance to predict the correct label for it.

Various types of neural network layers are commonly used for deep learning researches. Based on the type of the input, a layer can be designed to have parameters of different sizes and shapes, so that the layer can take maximal advantage of the input's spatial patterns. For example, if the input is an image, a ``convolutional layer'' can have a set of filters that acts on nearby pixels to generate output features, so that the spatial similarity between neighboring pixels is extracted. If the input is an English sentence, a ``recurrent layer'' can have a recursive structure within itself, so the relation between two adjacent words is represented. The choice of layer type is highly related to how the input data is formatted.

Multi-layer perceptron network (abbr. MLP network) is a common class of neural networks that consists of linear layers -- the most basic type of network layer. The parameters in a linear layer is simply a matrix: given an input of a 1-dimensional vector, the layer performs matrix multiplication on the vector and produces the result as another vector, a process equivalent to linear transformation. Often times, a fixed activation function acts on the result vector, so that non-linearity is added to the layer to make it more adjustable. The MLP network stacks several linear layers together, so that an input data vector is transformed into a feature vector, consisting of useful information for the machine to generate predictions. The intermediate linear layers are often called ``hidden layers'' of the network. Below is a figure illustrating an MLP network.

\begin{figure}[h!]
    \centering
    \includegraphics[width=0.7\textwidth]{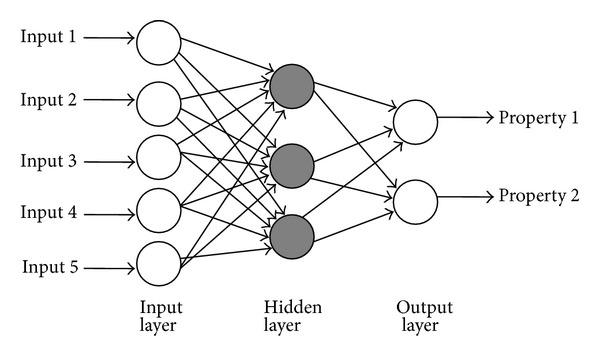}
    \caption{The concept of a MLP network}
    \label{fig:mlp1}
\end{figure}

MLP network is most suitable for input data that can be represented as 1-D vectors, in which different fields are not correlated with each other. As discussed in Section 3.2, the data we've designed highly resembles this structure, so MLP network is in fact the best choice of our task.

\subsection{Our Model Design}

Let's now go back to our task at hand: given a huge database of procurements, auditors need to first identify any suspicious procurements from the rest, then give each case a suitable label, so that punitive measures could be taken. So in fact, two models are needed for the above purposes: a binary model to distinguish between suspicious procurements and not suspicious ones, and a multiclass model to classify what kind of fraud is contained in a procurement already labeled as suspicious.

In fact, the two models can both be formulated as one single classification problem: the binary model needs to choose a prediction between class (1) this procurement is suspicious, or class (2) not suspicious, and the multiclass model needs to choose from a list of classes of possible fraud types. So in the process of designing our network, we can inherit our model from a given layer structure that has been proven to work with this type of inputs, and add a specific layer at the end for each model, in order to accommodate the different number of output classes. A ``softmax layer'' serves exactly this purpose: given a 1-dimensional tensor of many features, it can produce an output vector $v=(v_1,\ldots,v_n)$ of fixed length $n$, where each $v_i$ denotes the probability that this input should be classified as belonging to class $i$.

We decide to use the following structure for our network, provided in the example of training an MLP network on the MNIST dataset, which achieves state-of-the-art accuracy of $98.4\%$: \cite{mnistdata} \cite{mnistmlp}

\begin{figure}[h!]
    \centering
    \includegraphics[width=0.7\textwidth]{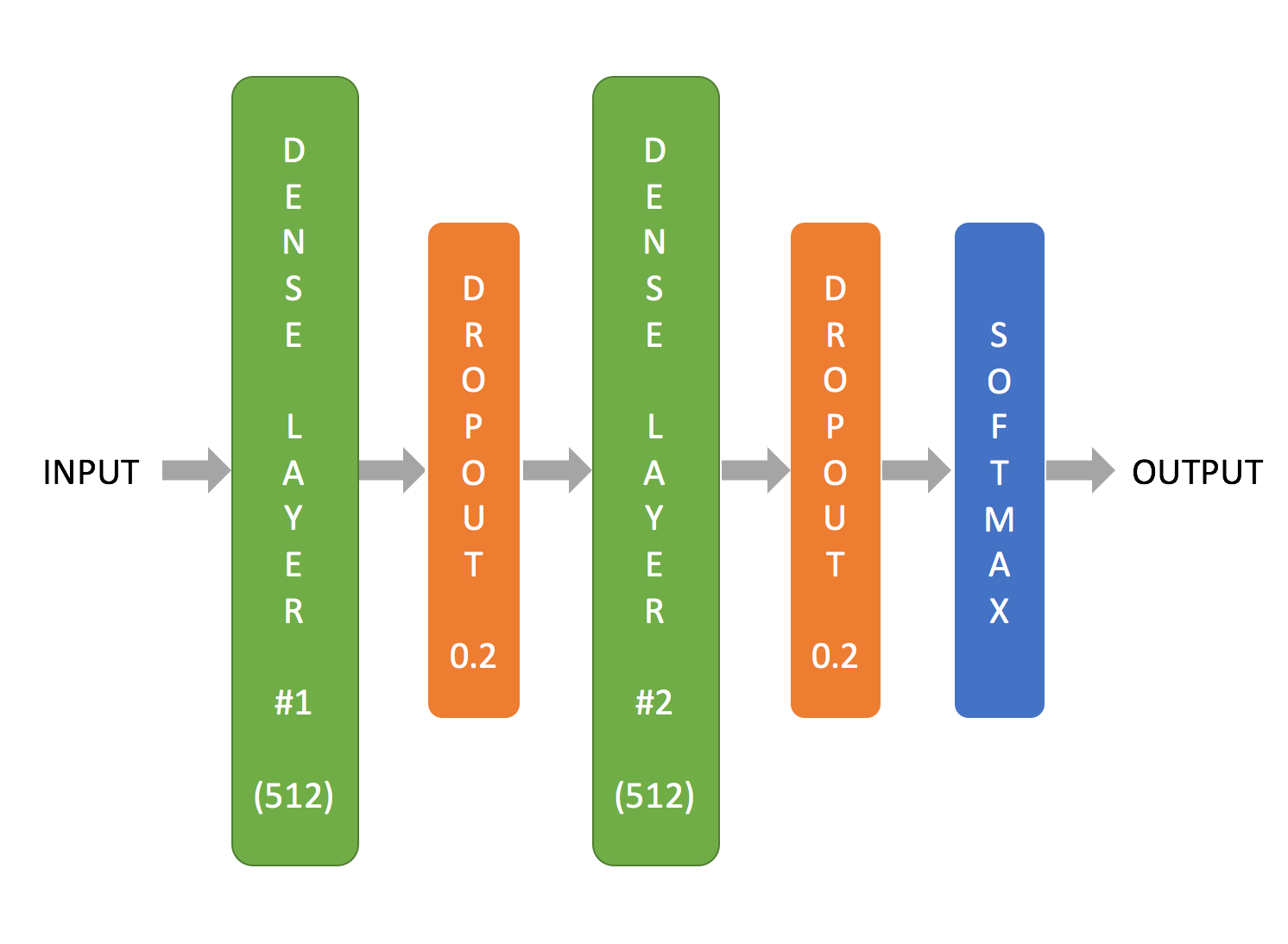}
    \caption{Structure of our MLP network}
    \label{fig:mlp2}
\end{figure}

In the above figure, the network contains two large, dense layers of 512 neurons, each followed by a dropout layer with ratio $d=0.2$. The purpose of the dropout layers is to prevent overfitting, which is a quite common problem in MLP networks. Next, the output layer is a softmax layer, as mentioned above, which fixes the number of output prediction classes, and should have different sizes for our two models. Note that although the two models share mainly the same structure, there are no relations between them in the training and predicting process, since each model is given a different ground truth label for an input data, and so the gradient descent process would happen in different directions, diverging the two models eventually to have very different parameters.

\section{Implementation, Testing and Results}

\subsection{Implementation Details}

In the actual coding, we choose to use two popular neural network libraries to implement our models: {\tt tensorflow} and {\tt keras}. First, {\tt tensorflow} serves as the backend of {\tt keras}, and provides strong GPU computing support when we train our networks on our large number of data samples. Second, {\tt keras} is well-known for its easy, clean abstractions of MLP network layers, which save us a lot of time when we code up, test, and optimize our neural network models.

In order to better initialize our networks and so achieve better performance, we decide to normalize all our input data so they fall into the range $[0,1]$. However, since the procurements are actually conducted at different scales, it is impossible to find a uniform standard of normalization for all data. So we make one pass through all samples, record the maximum and minimum values in each column, and perform linear normalization on each point by the formula:
\begin{equation}
normalized = \frac{original - min}{max - min}
\end{equation}

Our models are tested after training for 20 epochs, with a sample size of 50000, as specified in section 3.1. In addition, we implement 10-fold cross validation to get the average performance of our models on the whole dataset.

\subsection{Accuracy Results}

Below are the results of our binary model(left) and multiclass model (right), respectively:

\begin{equation}
\textup{
\begin{tabular}{|c|c|c|}
    \hline
    \textbf{Fold \#} & \textbf{Loss} & \textbf{Accuracy} \\\hline
    1 & $0.3417$ & $0.8297$ \\\hline
    2 & $0.2936$ & $0.8476$ \\\hline
    3 & $0.2755$ & $0.9273$ \\\hline
    4 & $0.3378$ & $0.8235$ \\\hline
    5 & $0.3343$ & $0.8251$ \\\hline
    6 & $0.3376$ & $0.7841$ \\\hline
    7 & $0.2985$ & $0.8460$ \\\hline
    8 & $0.3392$ & $0.8343$ \\\hline
    9 & $0.4273$ & $0.8333$ \\\hline
    10 & $0.3022$ & $0.8974$ \\\hline
    Average & $0.3288$ & \textbf{0.8448} \\\hline
\end{tabular}
\ \ \ \ 
\begin{tabular}{|c|c|c|}
    \hline
    \textbf{Fold \#} & \textbf{Loss} & \textbf{Accuracy} \\\hline
    1 & $0.0674$ & $0.9799$ \\\hline
    2 & $0.0896$ & $0.9778$ \\\hline
    3 & $0.0778$ & $0.9770$ \\\hline
    4 & $0.0551$ & $0.9770$ \\\hline
    5 & $0.0730$ & $0.9762$ \\\hline
    6 & $0.0734$ & $0.9778$ \\\hline
    7 & $0.0727$ & $0.9782$ \\\hline
    8 & $0.0800$ & $0.9782$ \\\hline
    9 & $0.0778$ & $0.9762$ \\\hline
    10 & $0.0859$ & $0.9807$ \\\hline
    Average & $0.0753$ & \textbf{0.9779} \\\hline
\end{tabular}
}
\end{equation}

Besides cross-validation, we've also extracted some actual data (different from the training samples above) to form a testing dataset. We choose 20 random rows from the dataset and record the prediction results of both models below. If we count the hits and misses, we can see that the accuracies shown here match quite well with the results obtained from above procedures.

\begin{figure}[h!]
    \centering
    \includegraphics[width=0.8\textwidth]{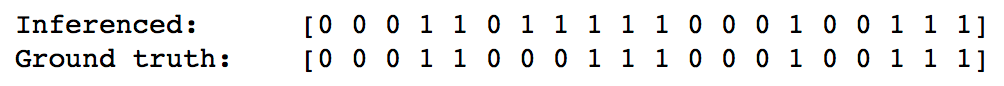}
    \caption{Binary model predictions}
    \label{fig:binary}
\end{figure}

\begin{figure}[h!]
    \centering
    \includegraphics[width=0.8\textwidth]{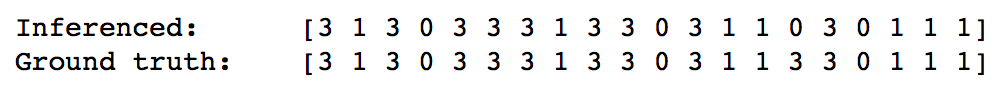}
    \caption{Multiclass model predictions}
    \label{fig:multiclass}
\end{figure}

Quite obviously, both models perform quite well in the training process. For the binary model, the result of $84\%$ accuracy is acceptable, based on the limited number of features. In an actual scenario, $16\%$ wrong prediction rate means that most of the procurement frauds would be caught by this model, which is far more efficient than traditional auditing techniques. Still, there could be ways to improve this result, such as adding more training features, or using more complicated network structures; it is possible to expect such a model to hit an accuracy of over $95\%$ if everything went smooth. And for the multiclass model, $98\%$ accuracy is actually the state-of-the-art performance of a multiclass MLP network -- the classification of fraud types is now working at its full power.

\section{Future Research \& Conclusions}

\subsection{Future Research}

The following subjects are to be studied.

\begin{itemize}[
    wide=0pt,
    labelwidth=1em,
    itemindent=0pt,
    leftmargin=\dimexpr\labelwidth+\labelsep\relax
]
\item Different information relevant to a single procurement might be stored in distinct databases. Such databases usually have different design methods and lengths of use, which makes data collection unnecessarily tedious. We suggest future researchers make "creating a more integrated database" one of their main targets.
\item Historical information related with fraud procurements is collected with high reliance on reports written manually. This fact significantly affects the number of procurements that could be used in machine learning. Considering the large-scale database needed in effective machine learning, more work should be done to increase the amount of cases that can be used to train the program.
\item Since raw data from databases have various formats, normalization measures need to be taken before we feed data to our models. However, due to the lack of consistency between columns of data, finding a good normalization method for all data is exceptionally hard. In our current experiments, we rely on some pre-determined assumptions on those data to construct different normalization formulae for different columns. It is urgent that a uniformed, concise and effective normalization method be worked out in future studies.
\end{itemize}

\subsection{Conclusions}

This research reveals both the probability of utilizing machine learning in traditionally manpower-consuming areas like fraud auditing and the difficulties that such utilization might face. The model testing part evidently shows that our model has a good performance in predicting the existence and further classification of fraudulence.

With the SAP database provided by SF Express, our artificial neural network exhibits decent calculating speed and accuracy, which lead us to the confidence of further applying machine learning in different types of audits.

A few incompatibilities still exist and further modifications of the model are required. Also, considering the fact that the database a company use directly decides the data to be inputted in training, the company itself should bring more uniformity to the database it use as well.

\section*{Acknowledgments}

We would like to offer special thanks to Zhijun Lin, Raymond Li, Peng Su, Chunlei Zhang, Jiansheng Fan and the auditing department of SF Express for all information and data concerning procurement fraud, as well as every single support they provided.

Also, we want to thank our reviewers and all of those who provided comments on prior drafts of this paper.

\bibliography{acl2016}
\bibliographystyle{unsrt}

\end{document}